\documentclass{article}\usepackage[hidelinks]{hyperref}

\usepackage{arxiv}

\usepackage[utf8]{inputenc} 
\usepackage[T1]{fontenc}    
\usepackage{hyperref}       
\usepackage{url}            
\usepackage{booktabs}       
\usepackage{amsfonts}       
\usepackage{nicefrac}       
\usepackage{microtype}      
\usepackage{lipsum}		
\usepackage{graphicx}
\usepackage[square,numbers]{natbib}
\usepackage{doi}

\usepackage{graphicx}
\usepackage[utf8]{inputenc}
\usepackage[english]{babel}
\usepackage{cancel}
\usepackage{amsmath}
\usepackage{url}
\usepackage{multirow}
\usepackage{array,multirow}
\usepackage{algorithm}
\usepackage{algorithmic}
\usepackage{graphicx}
\usepackage{verbatim}
\graphicspath{ {./figure/} }
\usepackage{babel}
\usepackage{xfrac}
\usepackage{dblfloatfix}
\usepackage{etoolbox}
\makeatletter
\patchcmd{\@makecaption}
  {\scshape}
  {}
  {}
  {}
\makeatother
\usepackage{xcolor}
\usepackage{tabularx}
\usepackage{amssymb}
\usepackage{amsthm}
\usepackage{array, boldline, makecell, booktabs}

\newtheorem{definition}{Definition}

\usepackage{multirow}
\usepackage{enumitem}
\usepackage{diagbox}
\usepackage{afterpage}

\usepackage{color,soul}
\usepackage{array,multirow}
\usepackage{caption}

\usepackage{microtype}
\usepackage{booktabs}
\usepackage[format=hang,font={small,bf}]{caption}
\usepackage[utf8]{inputenc}
\usepackage{color,soul}
\setul{0.5ex}{0.3ex}
\definecolor{Green}{rgb}{0,1,0}
\setulcolor{Green}
\usepackage[bottom]{footmisc}

\usepackage{nccmath}
\usepackage{mathtools}

\newcommand\blfootnote[1]{%
  \begingroup
  \renewcommand\thefootnote{}\footnote{#1}%
  \addtocounter{footnote}{-1}%
  \endgroup
}

\usepackage{fancyhdr}
\title{Evolving Reinforcement Learning Environment to Minimize Learner's Achievable Reward: An Application on Hardening Active Directory Systems}

\author{Diksha Goel, Aneta Neumann, Frank Neumann, Hung Nguyen, Mingyu Guo
\And
\normalfont{School of Computer and Mathematical Sciences}\\University of Adelaide, Australia\\
\{diksha.goel, aneta.neumann, frank.neumann, hung.nguyen, mingyu.guo\}@adelaide.edu.au}\vspace{-1in}

\vspace{-1in}

\begin{document}
\maketitle
\begin{abstract}
We study a Stackelberg game between one attacker and one defender in a configurable environment. The defender picks a specific environment configuration. The attacker observes the configuration and attacks via Reinforcement Learning (RL trained against the observed environment). The defender's goal is to find the environment with minimum achievable reward for the attacker. We apply Evolutionary Diversity Optimization (EDO) to generate diverse population of environments for training. Environments with clearly high rewards are killed off and replaced by new offsprings to avoid wasting training time. Diversity not only improves training quality but also fits well with our RL scenario: RL agents tend to improve gradually, so a slightly worse environment earlier on may become better later. We demonstrate the effectiveness of our approach by focusing on a specific application, Active Directory (AD). AD is the default security management system for Windows domain networks. AD environment describes an attack graph, where nodes represent computers/accounts/etc., and edges represent accesses. The attacker aims to find the best attack path to reach the highest-privilege node. The defender can change the graph by removing a limited number of edges (revoke accesses). Our approach generates better defensive plans than the existing approach and scales better.
\end{abstract}


\section{Introduction}
In an adversarial environment, the attacker and defender play against each other. The attacker intends to devise a technique to successfully carry out an attack, while the defender's objective is to protect the system from being compromised. In adversarial environments, the strategies and actions of the attacker and defender are interdependent and affect each other. This paper studies an attacker-defender Stackelberg game in configurable environment settings, where the defender (leader) tries various environment configurations to protect the system. In contrast, the attacker (follower) observes the environment configurations and designs an attacking policy using Reinforcement Learning (RL) to maximize their rewards. The defender aims to find an environmental configuration where the attacker’s attainable reward is minimum. We consider a specific application scenario, ``\textit{\textbf{hardening active directory systems}}'', to discuss the problem in detail.

Active Directory is a directory service developed by Microsoft for \textit{Windows domain networks}. It is designed to manage and secure network resources, such as user accounts, computers, printers, etc. 
AD is considered as a default security management system for Windows domain networks \cite{dias2002guide} and is widely adopted by a majority of large companies, due to which AD has become a popular target for cyber attackers. Microsoft reported that 90\% of Fortune 1000 companies use AD. As per the study by Enterprise Management Associates [2021], 50\% of the surveyed organizations have experienced an AD attack since 2019. 

AD attack graphs are widely used attack graph models among industrial practitioners and real-world attackers to model and analyze potential attack scenarios in an AD environment. The structure of AD describes an attack graph, with a node representing accounts/computers/etc., and directed edge $(i,j)$ indicating that an attacker can gain access to node $j$ from node $i$ via known exploits or existing accesses. \textsc{BloodHound} is a widely used tool for investigating AD graphs and identifies various attack paths in AD graph structures. \textsc{BloodHound} simulates an \textit{identity snowball attack}, in which an attacker begins from low-privileged account (gains access through phishing attack) and subsequently moves to other nodes with the goal of reaching the highest-privileged account,  \textsc{Domain Admin} (DA). \textsc{BloodHound} employs Dijkstra's shortest path algorithm to determine the path from entry node to DA. Given the sensitive nature of organizational AD and high number of cyber attacks targeting AD, security professionals are designing various solutions to defend AD. \textsc{BloodHound} is motivated by an academic paper \cite{dunagan2009heat}, where the authors designed a heuristic to selectively block some edges from the attack graph to disconnect the graph and prevent attackers from reaching DA. Notably, edge blocking in an AD environment is achieved by revoking accesses or implementing surveillance to prevent attackers from reaching the DA.

\textbf{\textit{This paper studies a Stackelberg game between an attacker and one defender on AD graphs, where {\textit{the attacker aims to design a strategy to maximize their chances of successfully reaching the DA, and the defender aims to design a strategy to minimize the attacker's success rate.}}}} The attacker in our model is strategic and adopts a proactive approach while performing the attack. Each edge in AD graph is associated with a failure rate, detection rate and success rate. The attacker starts the attack from one of the entry nodes and attempts to traverse an edge to reach new nodes. If the attacker fails to traverse the edge, the attacker tries another edge until gets detected, has tried all possibilities or reaches DA. If the attacker previously failed to pass through an edge, then they do not try this edge again. The strategic attacker maintains a set called \textbf{\textit{checkpoints}}, which are nodes that the attacker can use as starting points or continue an attack from. Initially, the checkpoint consists of only the entry nodes. The attacker tries various strategies to reach the DA and saves the information, such as the edges that the attacker fails to pass and that the attacker successfully passed. The endpoints of edges that the attacker successfully passed are added to checkpoints, and the attacker may try one of the unexplored edge starting from the nodes in checkpoints at any time during the attack. The optimal attacking policy is the one that maximizes attacker's chances of reaching DA without getting detected. Notably, it is essential to design a sophisticated attacker's policy, as we can not have an effective defense without having an accurate attacker’s policy. Furthermore, \textit{the defender's goal is to design a defensive policy to minimize the attacker's success rate.} The defender blocks a set of $k$ edges by increasing edge's failure rate from original to 100\%. In the literature, Goel et al. \cite{goel2022defending} studied the same AD model as ours. The authors trained the Neural Networks to solve the attacker problem and designed EDO based policy to address the defender problem. In contrast, we aim to propose a solution that scales better, approximates attacker's policy more accurately and generates better defensive plans than \cite{goel2022defending}. 

\noindent \textbf{Our contribution.} \textit{This paper aims to design the defender's policy to block a set of edges with the goal of minimizing the attacker's chances of successfully reaching the DA.}  For attacker problem, we propose a \textbf{\textit{Reinforcement Learning (RL)}} based policy to maximize attacker's chances of successfully reaching the DA (maximize attacker's achievable reward). We propose a \textbf{\textit{ Critic network assisted Evolutionary Diversity Optimization (C-EDO)}} based defensive policy to find defensive plan configurations that minimize the attacker's success rate. The attacker's problem of maximizing their chances of successfully reaching DA can be modelled as a Markov Decision Process (MDP). We propose RL based policy to approximate the attacker's problem and use \textit{Proximal Policy Optimization} RL algorithm, an \textit{Actor-Critic based approach} to train the attacker policy. The RL agent interacts with ``\textit{multiple environments}'' simultaneously by suggesting actions with the goal of maximizing the overall reward. We propose Critic network assisted Evolutionary Diversity Optimization based policy to solve the defender problem. C-EDO generates numerous environment configurations (defensive plans). 
Our approach uses the trained RL critic network to estimate the fitness of environment configurations. The defender continuously monitors the RL training process and after regular intervals, defender uses the trained critic network to evaluate the current configurations and uses C-EDO to generate better ones. The attacker and defender play against each other in parallel. 

Overall, the defender's C-EDO generates numerous high-quality, diverse environment configurations worth learning for the RL agent to train a better attacking policy. 
We train our attacker's policy using actor-critic based algorithm, so we inherently have a critic network that gives us a reasonable estimation of the state values and can be used as a fitness function for defender's C-EDO. 
In this way, these two problems are interconnected, and the solution strategies complement each other. We conduct extensive experimental analysis and compare our results with an existing approach \cite{goel2022defending}. Our results demonstrate that 1) Proposed approach is \textbf{\textit{scalable}} to r4000\footnote{r4000 AD graph is an AD graph containing 4000 computers.} AD graph, for which the existing approach fails to scale; 2) Proposed approach is  \textbf{\textit{highly effective}}. For r1000 AD graph, the best defense from our proposed approach has an average of 41.01\% chances of success, significantly less than the other approaches; 3) Proposed RL based policy \textbf{\textit{approximates attacker's problem more accurately}}. For r1000  AD graph, the attacker's average success rate is increased by 0.63\% using our attacking policy; 4) Proposed approach \textbf{\textit{generates better defense}}. For r1000  AD graph, the attacker's average success rate is reduced by 1.76\% under our proposed defense. 

Our proposed approach achieves better results than \cite{goel2022defending} as we have used RL to approximate the attacker's problem. NNDP attacker's policy \cite{goel2022defending} trains the model against one defensive plan at a time, due to which it forgets the previous plan. This way, it keeps learning and forgetting the plans. However, we train our RL based attacker's policy against multiple defensive plans at a time, due to which it learns shared experience and performs better. For RL agent, diverse environment configurations are only different in the "opening games", whereas the "end games" or "mid games" are likely to be similar across different environments. The similarity in later stages can be utilized in parallel training, where the agent is trained against multiple environments simultaneously and gains shared experience, leading to faster convergence and improved performance. Besides, NNDP approach is value iteration-based RL algorithm, whereas our approach is policy iteration-based RL algorithm. In general, the policy iteration-based algorithms converge faster than value iteration-based algorithms \cite{kaelbling1996reinforcement}, which is another reason for the superior performance of our approach. Our experimental results further support our arguments. 

\section{Related Work}
\noindent \textbf{Active Directory.} Goel et al. \cite{goel2022defending} studied the same Stackelberg game model between an attacker and defender on AD graphs as ours. The authors used neural networks to address the attacker problem and EDO for the defender's problem. However, their approach is not scalable to large AD graphs. Our proposed approach is scalable to large AD graphs and can approximate the attacker's problem more effectively. Guo et al. \cite{guo2022practical} studied a different model for defending AD graphs, in which the defender aims to maximize the expected path length of attacker. The authors proposed fixed-parameter algorithm and graph convolutional neural network based approaches to discover edges to be blocked. Guo et al. \cite{guo2022scalable} proposed various scalable algorithms for defending AD attack graphs. However, the problem settings that they considered are different from ours. The authors proposed a tree decomposition based dynamic program and RL based approaches that are scalable to large AD graphs. Ngo et al. \cite{quang} designed near-optimal policy for the placement of honeypots on computer nodes when the AD graph is dynamically changing; however, the authors studied identifying blocking nodes, whereas we focused on blocking edges. Zhang et al. \cite{Yumeng23:Near} proposed a scalable double oracle algorithm and compared their solution against various industry solutions; the authors considered different problem settings than ours.

\noindent \textbf{Evolutionary Diversity Optimization. }
Evolutionary Diversity Optimization discovers high quality maximally different solutions. EDO was first studied by Ulrich et al. \cite{ulrich2010integrating} and has attained significant attention in the evolutionary computation community. Huang et al. \cite{huang2023co} designed a co-evolutionary approach to improve the searchability and convergence of competitive swarm optimizer. 
Bossek et al. \cite{bossek2021evolutionary} investigated EDO for computing a diverse set of solutions for minimum spanning tree problem. 
Do et al. \cite{do2022analysis} investigated EDO for permutation problem. Nikfarjam et al. \cite{nikfarjam2021entropy} designed an entropy-based evolutionary approach to compute diverse solutions for travelling salesperson problem. 

\section{Problem Description}
Active directory graph can be represented as a directed graph $G = (V, E)$, where $V$ is nodes set, and $E$ is edges set. The highest privilege accounts in AD are called \textsc{Domain Admin} (DA). This paper considers a two-player Stackelberg game between one defender and one attacker to defend AD graphs. There are $s$ entry nodes. The attacker can start from one of the entry nodes and aims to design a strategy to maximize their chances of successfully reaching DA. The defender seeks to block a set of edges so as to minimize the attacker's success rate. The defender has a limited budget and can only block $k$ edges. Not all edges are blockable; therefore, the attacker can only block `\textit{blockable}' edges. Edge blocking is costly and requires efforts (auditing access logs) to safely remove an edge; due to this, we cannot remove too many edges and have assumed a limited budget. In our model, every edge in the AD graph has a detection rate, failure rate and success rate. \textit{Detection rate} $p_{d(e)}$ indicates that if an attacker gets detected, the attack terminates immediately. \textit{Failure rate} $p_{f(e)}$ indicates that if the attacker fails, then they are unable to traverse edge $e$ successfully (attacker might have to enter a correct password to pass through edge). In this case, the attack does not terminate, and the attacker can continue from other checkpoints by trying unexplored edges. \textit{Success rate} $p_{s(e)}$ denotes the attacker's likelihood of successfully passing through edge $e$ and is calculated as $p_{s(e)} = 1- p_{f(e)} - p_{d(e)}$. The strategic attacker starts from one of the entry nodes and tries unexplored edges in order to reach DA. At any time during the attack, the checkpoint indicates the set of nodes that the attacker has control of and can continue the attack from (in case the attacker fails to pass through the edge). Goel et al. \cite{goel2022defending} proved that computing defender's and attacker's optimal policy (and value) is $\#$P-hard. Therefore, to approximate the attacker problem, we design a reinforcement learning based policy where the agent learns from multiple environment configurations (defensive plans) at a time, in turn accelerating the training process. We propose a critic network assisted evolutionary diversity optimization policy to address the defender problem.

\section{Proposed Approach}
This section describes our proposed approach for defending active directory graphs. We first discuss our proposed pre-processing procedure that converts the original AD graph to a smaller graph. We then discuss our proposed reinforcement learning based attacker's policy and critic network assisted evolutionary diversity optimization based defender's policy. Later, we describe our overall attacker-defender approach.

\subsection{Pre-processing AD graphs}
We first pre-process our AD graph by exploiting its structural features to get a smaller graph. In an AD graph, \textit{Splitting nodes} are the nodes with multiple edges originating from them. \textit{Entry nodes} are the starting points from where the attacker can initiate an attack. The set of splitting nodes and entry nodes is represented by \textsc{Split} and \textsc{Entry}, respectively.

\begin{definition}
\label{def_NSP}
\noindent \textbf{Non-Splitting Path (NSP).}  Non-Splitting Path NSP $(i, j)$ is a path from node $i$ to $j$, where $j$ is the only successor of node $i$; then iteratively moves to the sole successor of $j$, until DA or splitting node is encountered \cite{guo2022scalable}.
\end{definition}
\begin{equation*}
NSP = \{NSP(i,j)|\, i \in  \textsc{Split} \cup \textsc{Entry}, j \in \textsc{Successors}(i) \} 
\end{equation*}
If at least one of the edges on an NSP is blockable, only then we say that the NSP is blockable.

\begin{definition}
\label{def_bw}
\noindent \textbf{Block-worthy edge (bw).}  Any blockable edge farthest away from node $i$ on $NSP(i,j)$ is known as block-worthy edge $bw(i,j)$. The block-worthy edge set is defined as:
\end{definition}
\begin{equation*}
BW = \{bw(i,j)|\, i \in \textsc{Split}\, \cup \,\textsc{Entry}, j \in \textsc{Successors}(i)\}
\end{equation*}
A block-worthy edge can be shared among two NSPs. We only spent one budget unit on blocking $NSP(i,j)$. 
\textit{If the original graph contains $n$ nodes and $m$ edges; after pre-processing, the resulting graph consists of ($|\textsc{Entry}|+|\textsc{Split}| + 1)$ nodes and $|NSP|$ edges.}

\subsection{Attacker Policy: Reinforcement Learning }
The attacker's goal is to design a strategy that maximizes their chances of successfully reaching the DA. We describe the attacker's problem as a Markov Decision Process and propose a reinforcement learning based policy to address the attacker's problem. Our proposed RL based attacker policy uses \textit{Proximal Policy Optimization (PPO)} algorithm, an Actor-Critic based approach, to train the RL agent. 
\textit{{We train our RL agent in parallel against multiple instances of environment configurations at a time, and each environment contains a defensive plan from the defender}}. 
This section discusses our proposed RL based attacker's policy in detail.

\subsubsection{Environment}
We model the attacker's problem of designing a policy to maximize their chances of successfully reaching the DA as MDP. We call the attacker's MDP $M = (S, A, R, T)$  as an environment, where $S$ denotes the state space, $A$ denotes the action space, $R$ is the reward function, and $T$ represents the transition function. The description of the environment is discussed below.\\

\noindent \textbf{\textit{State space (S):}} State space $S$ is a finite set of attacker's states, and state $`s$' is a vector of size equal to the number of NSPs in AD graph. Each coordinate in state $s$ represents one NSP, and the status of each NSP is either `S', `F' or `?'. We represent the attacker state $s$ as:
\begin{equation}
\label{attacker_state}
\text{Attacker state s}  = {\underbrace{<S, F, ?, . \,. \,. \,. \,., F, ?, S, ?>}_\text{Length = Number of NSPs}}
\end{equation}
where the status `S' indicates that the attacker has successfully reached the other end of NSP after attempting it, `F' signifies that the attacker failed to reach the other end despite trying, and `?' signifies that the attacker has not tried this NSP yet.

Given a state $s$, the attacker explores one of the NSPs with status  `?' and the status of tried NSP changes to either `S' or `F' \footnote{Status of some other NSPs may also change, in case the destination is already reached, or joint block-worthy edge is failed.}, and the attacker reaches a new state. The attacker continues to explore one of the unexplored NSPs at a time to reach a new state till the attacker reaches DA or gets detected. At any time $t$ during the attack, the attacker's current state $s_t$ acts as a  knowledge base and conveys the following information: the set of NSPs that the attacker has control of (NSPs with status `S'), NSPs that the attacker has failed on (NSPs with status `F') and NSPs that the attacker can try in future (NSPs with status `?'). The attacker has two base states: 1) When the attacker reaches DA, the attack is successful and terminates; 2) When the attacker is not able to reach the DA; the reason can be that they got detected or has tried all possible NSPs, and there is no option left to explore; in this case, the attack fails and ends.

\noindent \textbf{\textit{Action space (A):}} Action space $A$ is the action set available for state $s$, which are the outgoing NSPs from the successful NSPs in state $s$. The attacker's action space is discrete. Action $a$ is one of the NSPs from the action space of state $s$ and indicates that the attacker tries this NSP to reach the DA. 

\noindent \textbf{\textit{Reward function (R):}} The reward $r(s, a)$ for state $s$ on performing an action $a$ is $1$ if the attacker is able to reach the DA without getting detected. Otherwise, the reward is $0$.

\noindent \textbf{\textit{Transition function (T):}} For a given state $s$ and action $a$, the transition function performs action $a$ on state $s$, and may have a set of future states. Each future state is associated with its transition probability, and one state is selected as the next state (weighted by its transition probability). 


\subsubsection{Policy training}
We propose to utilize \textit{Proximal Policy Optimization (PPO)} RL algorithm to train the attacker's policy so as to maximize attacker's success rate. PPO follows an \textit{{Actor-Critic approach}} that exploits the advantages of policy based and value based approaches while eliminating their disadvantages. In this approach, the policy and value networks help each other improve. In our approach, we train two networks: actor network and critic network. \textbf{\textit{Actor network}}, also referred to as policy network, takes the current state $s$ as input and outputs the action $a$ to be performed on $s$. Actor network updates the policy parameters in the direction implied by the critic network. \textbf{\textit{Critic network}}, also known as value network, takes the state as input and outputs the value of state. \textit{For an attacker problem, value of the state is the attacker's success rate.}

Our RL agent uses the actor-critic based PPO algorithm to train the attacker's policy by interacting with the environment (each environment is associated with a configuration, i.e., defensive plan). The RL agent does not possess any prior knowledge of the environment. Instead of training the RL agent against a single environment, we train the agent against \textit{\textbf{multiple parallel environments}}.  Each environment is initialized with the attacker's initial state (defensive plan is converted to obtain attacker's initial state). The following process is executed in all environments simultaneously. At each timestep $t$ of an episode, the RL agent receives state $s_t$. The trained actor network issues an action $a_t$ to be performed on the current state $s_t$ and action $a_t$ is sent to the environment. The environment performs action $a_t$ on state $s_t$ and reaches a new state $s_{t+1}$ (following the transition function), and receives a reward $r_{t+1}$ (following the reward function). The process repeats until we reach the base state, i.e., the attacker reaches DA or gets detected. In this manner, we obtain a sequence of states, actions, and rewards that terminates at the base state. The designed policy intends to maximize the total reward received during an episode. The final reward is the attacker's success rate. \textbf{\textit{This way, the attacker designs RL based policy to maximize their achievable reward (success rate).}}

\subsection{Defender Policy: Critic Network Assisted Evolutionary Diversity Optimization}
The defender's goal is to block $k$ block-worthy edges to minimize the attacker's chances of successfully reaching the DA. We propose a \textit{\textbf{Critic network assisted Evolutionary Diversity Optimization (C-EDO)}} based defensive policy that computes high quality and diverse environmental configurations (defensive plan). We aim to identify the valuable environments, i.e., the environment configurations that correspond to potentially good defense. Our main idea is to let the RL agent play against the environment configuration and if, after training for some time, the configuration proves to be unfavourable for the defender (i.e., the attacker has a high success rate against the configuration), we discard this environment configuration. We do not waste our computational effort on this environment and allocate our limited computational resources to other challenging environments. The high-quality and diverse characteristics of environments enhance the  accuracy of modelling the attacker. The trained RL critic network serves as a fitness function for the defender's C-EDO. For every environmental configuration, the fitness function is used to evaluate that environment, i.e., calculate the attacker's success chances against that configuration (defensive plan). The defender only blocks block-worthy edges to generate environment configuration. The defender's environment configuration/defensive plan can be represented as:

\begin{equation}
\label{def_state_vec}
\text{Environment configuration} \,\, = \,\,< N, B, \,.\, .\, .\, ,\,B, N, B> 
\end{equation}
where `B' denotes the blocked edges and `N' denotes the non-blocked edges. Notably, length of the configuration (defensive plan) is equal to the number of block-worthy edges in AD graph.

\begin{algorithm*}[t!]
\caption{Critic network assisted Evolutionary Diversity Optimization based Defender's Policy}
 \label{edo_algo}
 \begin{algorithmic}[1]
 \STATE Initialise population P with $\mu$ environment configurations
 \STATE Select individual $p'$ uniformly at random from $P$ and create an offspring $p'_{new}$ by mutation or crossover
 \STATE If $(OPT - 0.1) \leq fitness(p'_{new}) \leq (OPT + 0.1)$, add $p'_{new}$ to $P$
 \STATE If $|P| = \mu + 1$, remove  individual $r$ from $P$ that contributes least to diversity, i.e., minimum $SortedDiver(C(bw)\backslash{p_r})$
 \STATE Repeat steps 2 to 4 till the termination condition is met
\end{algorithmic}
\end{algorithm*}

Algorithm \ref{edo_algo} outlines the defender's policy. The process starts by generating arbitrary population $P$ of $\mu$ configurations as shown in Eq. (\ref{def_state_vec}). In every configuration, the total number of $Bs$ is $k$ (defender's budget). To create new offspring (environmental configuration), we perform mutation or crossover operation, each with $0.5$ probability on the randomly selected configuration $p'$ from the population $P$. We randomly select a variable $x$ from a Poisson distribution with mean value of 1. For \textit{\textbf{Mutation}}, we select an individual $p'$ randomly from $P$ and flip $x$ $Bs$ to $Ns$ and $x$ $Ns$ to $Bs$. For \textit{\textbf{Crossover}}, we randomly select two individual $p'$ and  $p''$ from $P$ and look for $x$ indices such that $p'$ has $N$ and $p''$ has $B$ on those indices. Now, for these indices, we change $Ns$ to $Bs$ in $p'$ and $Bs$ to $Ns$ in $p''$. Similarly, we look for $x$ indices where $p'$ has $B$ and $p''$ has $N$ on those indices and change $Bs$ to $Ns$ in $p'$ and $Ns$ to $Bs$ in $p''$. The mutation and crossover operation make sure that exactly $k$ edges are blocked in the environmental configuration. We add the newly created offspring to the population only if its fitness score lies within the range of $(OPT \pm 0.1)$; otherwise, the offspring is rejected even though it is beneficial for diversity. If the new offspring is added to the population, we aim to maximize the blocked edge diversity and remove the individual that contributes least to the diversity. \\


\noindent \textbf{Maximizing blocked edges diversity.} We define the diversity measure as ``\textit{all block-worthy edges are equally blocked}". Our proposed diversity measure aims to maximize the diversity of blocked edges in the environment configurations. It calculates how often each block-worthy edge is blocked in the configuration population, with the aim of making this frequency equal. When a new offspring is created using mutation or crossover, the offspring is added to the population only if its fitness score is close to the best fitness score and rejects the individual that contributes least to the diversity. Let us assume there are $\mu$ configurations (we call these configurations as individuals) in $P$. Each individual $p_i$ can be described as:

\begin{equation*}
  p_i =  \big((B/N, bw_1), (B/N, bw_2), ..., (B/N,bw_{|BW|})\big)
\end{equation*}
where B/N denotes the status of block-worthy edge; `B' indicates blocked, `N' indicates non-blocked, and $i \in \{1, . . .,\mu\}$. We then compute the block-worthy edge count vector $C(bw)$, which is defined as ``for each block-worthy edge $bw_j, \,\, j \in \{1, ..., |BW|\}$, the number of individuals who have blocked $bw_j$ edge''.
\begin{equation*}
   C(bw) = (c(bw_1), c(bw_2), ..., c(bw_{|BW|}))
\end{equation*}
Where $c(bw_1)$ denotes the total individuals out of $\mu$ who have blocked $bw_1$ edge. For each individual $p_i$, we then determine the vector $Diver(C(bw)\backslash{p_i})$, which computes the diversity of the population without individual $p_i$ as:
\begin{equation*}
\label{diversity_eq}
  Diver(C(bw)\backslash{p_i}) = C(bw) - p_i
\end{equation*}
Our goal is to maximize the blocked edge diversity; therefore, we calculate $SortedDiver(C(bw)\backslash{p_i})$ as:
\begin{flalign*}
\text{\textit{SortedDiver}} (C(bw)\backslash{p_i})=  sort\Big(Diver(C(bw)\backslash{p_i})\Big)
\end{flalign*}

\begin{figure}[h!]
\centering
 \includegraphics[width=0.5\paperwidth]{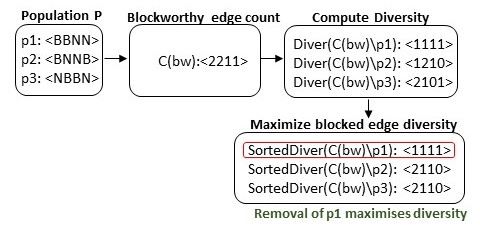}
 \caption{Example of maximizing blocked edge diversity.}
 \label{fig:edge_diversity_eg}
\end{figure}

To maximize the diversity of blocked edges, our goal is to minimize the $SortedDiver(C(bw)\backslash{p_i})$ in lexicographic order where sorting is carried out in descending order. The individual $l$ with minimum $SortedDiver(C(bw)\backslash{p_l})$ is the one, removal of which maximizes the diversity. Therefore, the individual $l$ is removed from the population, if its removal maximizes the diversity and its fitness score is not close to the best. In special case, when the newly created offspring has the best fitness score, we add it to $P$ (even though it is worst in terms of diversity) and the individual with the lowest fitness score is discarded. Using this process, the defender creates diverse environment configurations. Figure \ref{fig:edge_diversity_eg} presents an example of maximizing the blocked edge diversity in population.

\subsection{Attacker-Defender Overall Approach}
The defender employs C-EDO to generate high-quality and diverse environment configurations, each containing a defensive plan worth learning for the attacker policy. The attacker uses an actor-critic based RL algorithm to train their policy against the defender's environment configurations. The attacker's trained RL critic network serves as a fitness function for the defender C-EDO. The attacker first converts the defensive plans in environment configurations (Eq. (\ref{def_state_vec})) to the attacker's initial state\footnote{In attacker's state, the status of NSPs corresponding to the blocked block-worthy edges in the defender's environment configuration is changed from `?' to `F' to obtain attacker's initial state.} (Eq. \ref{attacker_state})). The RL agent then interacts and learns from the environments in parallel by issuing actions according to the trained policy. The environments perform the action and return a new state and reward. The quality of the trained policy is determined based on the total reward collected by the agent during an episode. Initially, the policy is not trained, and the action suggested might result in low rewards, but with training, it results in high rewards (attacker's success rate). In this way, the RL agent trains the policy to maximize the reward. 

\begin{figure}[h!]
\centering
\includegraphics[width=9cm,height=4cm]{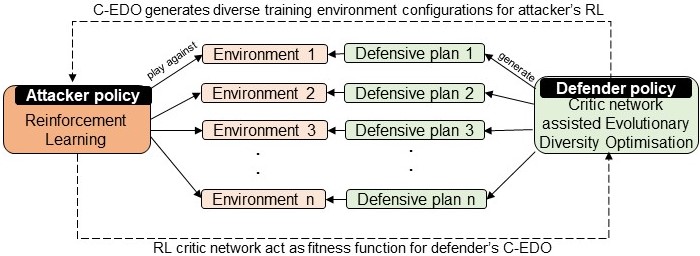}
 \caption{Overall attacker-defender proposed approach.}
 \label{fig:overall_approach}
\end{figure}

The defender process continuously monitors the RL training process and, after every regular interval, uses the trained critic network to evaluate the current set of environments. The defender discards those environments that are good for the attacker and replaces them with better ones. This way, at the beginning of every episode, the RL agent checks if the defensive plan configuration corresponding to each environment is changed or not. If changed, the attacker initializes the environment with the new defense configuration and starts training the agent against it. In this manner, the attacker and defender play against in parallel, where the attacker's RL process continuously learns, and the defender process evaluates the current environment configurations and generates better ones. C-EDO's diversity characteristic helps RL not get stuck in the local optimum. 
The RL agent experiences differences in environmental configurations only in the early stages, but the later stages tend to be similar across environments. We took advantage of this similarity and trained the agent on multiple environments in parallel, due to which our agent gained shared experience, resulting in faster convergence and enhanced performance. Overall, the trained attacker's RL policy improves as the defender generates better environmental configurations using C-EDO. The defender's policy generates better environments as the attacker's critic network is trained. This way, these two processes assist each other to perform better. Figure \ref{fig:overall_approach} illustrates our overall proposed approach.
\section{Experimental Results}
We executed experiments on a high-performance cluster server with Intel Gold 6148/6248 CPUs, utilizing 1 CPU and 20 cores for each trial. We used OpenAI Gym \cite{brockman2016openai} to implement the RL environments and trained the RL model using the PPO algorithm from the \textsc{Tianshou} library \cite{weng2021tianshou}. \textit{Success rate/Chances of success} indicates the attacker's probability of reaching the DA without getting detected when the defender has blocked certain edges.

\subsection{Dataset}
Real-world AD graphs are highly sensitive; therefore, we used \textsc{BloodHound} team’s synthetic graph generator \textsc{DBCreator} to generate synthetic AD graphs. We generate AD graphs of four sizes, i.e., r500, r1000, r2000 and r4000, where 500, 1000, 2000 and 4000 indicate the number of computers in the AD graph. r500 AD graph contains 1493 nodes and 3456 edges; r1000 AD graph contains 2996 nodes and 8814 edges; r2000 AD graph contains 5997 nodes and 18795 edges; r4000 AD graph contains 12001 nodes and  45780 edges. Our experiments consider only 3 specific kinds of edges, by default present in \textsc{BloodHound}: \textsc{HasSession}, \textsc{MemberOf} and \textsc{AdminTo}. To set the attacker's starting nodes, we first find 40 faraway nodes from DA and then arbitrarily set 20 nodes from them as the starting nodes. Each edge $e$ is blockable with a probability =  $\frac{\text{Min \#hops between e and DA}}{\text{Max \#hops between any e and DA}}$. This way, the edges farthest from the DA, i.e., less significant edges, are more likely to be blocked. It is challenging to perform computations on a large AD graph; therefore, we pre-process it to obtain a smaller graph. Pre-processing includes merging all DA into 1 DA, removing nodes and edges irrelevant for the attacker (outgoing edges from DA,  all incoming edges to the entry nodes and the nodes without any incoming edges). We also consider an NSP (Definition \ref{def_NSP}) as one edge. 
Furthermore, to investigate the relationship between the failure rate $p_{f(e)}$ and detection rate $p_{d(e)}$ of an edge $e$ on attacker's chances of success, we use three different distributions: Independent distribution, Positive correlation, and Negative correlation. In \textit{independent distribution}, we set the values of $p_{d(e)}$ and $p_{f(e)}$  as:
\begin{equation*}
p_{d(e)}, p_{f(e)}  = \text{Independent uniform (0, 0.2)}
\end{equation*}

\noindent In \textit{positive correlation}, we set the values of $p_{d(e)}$ and $p_{f(e)}$ as:
\begin{equation*}
p_{d(e)}, p_{f(e)} = \mathcal{N}  (\mu, \Sigma)
\end{equation*}
where
\begin{equation*}
\mu = [0.1, 0.1] \text{\,\,and\,\,} \Sigma = [[0.05^2, 0.5\times0.05^2], [0.5 \times 0.05^2, 0.05^2]]
\end{equation*}
Here, $\mathcal{N}$ represents the multivariate normal distribution, $\mu$ represents the mean, and $\Sigma$ represents the covariance matrix.

\noindent In \textit{negative correlation}, we set the values of $p_{d(e)}$ and $p_{f(e)}$ as:

\begin{equation*}
p_{d(e)}, p_{f(e)} = \mathcal{N}  (\mu, \Sigma)
\end{equation*}
where
\begin{equation*}
\mu = [0.1, 0.1] \text{\,\,and\,\,} \Sigma = [[0.05^2, -0.5\times0.05^2], [-0.5 \times 0.05^2, 0.05^2]]
\end{equation*}

\subsection{Training Parameters}
\textit{\textbf{Reinforcement Learning:}}
We used a simple multi-layer perceptron neural network to implement the actor and critic network. 
The hidden layer size is 128 for r500 and r1000 AD graphs,  and 256 for r2000 and r4000 AD graphs. The parameters are trained using adam optimizer, learning rate of $1e^{-4}$ and batch size of $800$ states. For PPO-specific hyper-parameters, we used the standard hyper-parameters as specified in the original paper \cite{schulman2017proximal}. We created $20$ environments. For experimental setup 1, we parallelly train the RL policy for a total of $700$ epochs (1200 epochs for r2000 and r4000 AD graph) on $20$ environments. After every $20$ training epochs, the defender evaluates and resets the environments. When the terminating condition is met (number of epochs), the defender chooses the defensive plan with the lowest attacker success rate as their \textit{best environment configuration}. For experimental setup 2 and 3, we train the RL policy for $150$ epochs on $20$ environments parallelly. The trained attacker's policy is then simulated on the best environment configuration for $5000$ episodes, and the average reward over $5000$ episodes is the attacker's success rate. 

\noindent \textit{\textbf{Critic network assisted Evolutionary Diversity Optimization:}} The defender can only block $5$ edges. In $20000$ iterations, defender creates a population of $20$ environment configurations (defense).

\begin{table*}[t!] 
\caption{Comparison of attacker's chances of success under various attacker-defender policies (smaller number represents better performance). Results show that the proposed C-EDO defensive policy leads to the best defense. Also, our RL+C-EDO policy scales to r4000 graphs, for which NNDP-EDO \cite{goel2022defending} failed to scale. }
\label{r1000}
\renewcommand{\arraystretch}{1.1}
\centering 
\begin{tabular}{p{.8cm}p{3.3cm}p{1.5cm}p{1cm}p{1cm}p{1.2cm}p{1.5cm}p{1cm}p{1.2cm}} \hlineB{2}
\multicolumn{1}{c}{\textbf{{}}}& \multicolumn{1}{c}{\textbf{{}}}&  \multicolumn{4}{c}{\textbf{{Chances of success}}} & \multicolumn{3}{c}{\textbf{{Time (hour)}}} \\  \cmidrule(lr){3-6}  \cmidrule(lr){7-9}

\textbf{Graph} & \textbf{Policy} & \textbf{Independent} & \textbf{Positive} & \textbf{Negative} &  \textbf{Average} &\textbf{Independent} & \textbf{Positive} & \textbf{Negative}\\ \hlineB{2} 
      & RL+C-EDO (Proposed) & \textbf{40.16\%} & \textbf{41.36\%} & \textbf{41.51\% }& \textbf{41.01\%} & 43.82 & 44.73  &  47.27\\
r1000 & RL+EC & 41.69\% & 48.45\% & 42.89\% & 44.34\% & 44.43 & 43.88  & 48.43\\
      & RL+Greedy & 56.45\%  & 47.86\% & 47.88\% & 50.73\% & 43.32 & 44.95 & 47.38 \\\hlineB{2} 
      & RL+C-EDO (Proposed) & \textbf{25.09\%} & \textbf{32.45\%} & \textbf{30.44\%} & \textbf{29.33\%} & 112.74 & 118.58  &  119.53\\
r2000 & RL+EC & 28.13\% & 35.51\% & 37.28\% & 33.64\% & 114.22 &  112.57 & 118.51\\
      & RL+Greedy & 33.43\%  & 34.30\% & 40.06\% & 35.93\% & 113.84 & 116.43 &  113.76\\\hlineB{2} 
      & RL+C-EDO (Proposed) & \textbf{22.02\%} & \textbf{17.29\%} & \textbf{21.81\%} & \textbf{20.37\% }& 127.54 & 121.11  &  137.73\\
r4000 & RL+EC & 22.78\% & 21.87\% & 24.71\% & 23.12\% & 132.55  &  120.67 & 135.31\\
      & RL+Greedy & 25.16\%  & 24.18\% & 22.34\% & 23.89\% & 125.61 & 120.73 & 127.29 \\\hlineB{2} 
\end{tabular} 
\end{table*}

\subsection{Experimental Setup 1}
In this experimental setup, we determine the effectiveness of our overall proposed approach. 

\subsubsection{\textbf{Baseline}}
We combine the RL based attacker's policy with various defender's policies to compare the effectiveness of our proposed defensive policy\blfootnote{* indicates that with our general parameter settings, RL policy results were slightly bad than baseline. Therefore, we train the RL policy for 300 epochs instead of 150 epochs. Given enough time, the RL policy outperforms the baseline.}. 
\begin{itemize}[leftmargin=*]
    \item \textit{\textbf{RL+C-EDO (proposed)}}: In this approach, RL is utilized as attacker's strategy, whereas C-EDO is used as defender's policy. In C-EDO, the defender rejects those environment configurations that contribute least to diversity. 
    \item \textit{\textbf{RL+EC}}: In RL+EC approach, RL is utilized as attacker's policy, and Evolutionary Computation (EC) is used as defender's policy. In EC, the configurations with the lowest fitness score are discarded.
    \item \textit{\textbf{RL+Greedy}}: In RL+Greedy approach, RL is utilized as attacker's strategy. The defender uses a greedy technique to generate environment configurations. The defender greedily blocks one edge that minimizes attacker's success rate. This way, the defender iteratively discovers $k$ edges to be blocked.
\end{itemize}

\subsubsection{\textbf{Results}} We perform experiments on $r1000$, $r2000$ and $r4000$ AD graphs. We first train the RL based attacker's policy on the environment configurations (defensive plan) generated by the defender's policies, i.e., C-EDO, EC and Greedy. We then test the effectiveness of the attacker's policy against the defender's best environmental configurations. We report the average reward (success rate) by simulating the attacker's strategy on the best environment for $5000$ episodes. For each AD graph, we perform experiments on $5$ seeds from $0$ to $4$, and report the average success rate over $5$ seeds. Table \ref{r1000} reports the results obtained for r1000, r2000 and r4000 AD graph. For r1000 AD graphs, results from the table show that under independent distribution, the attacker's chances of success under C-EDO based defensive policy is 40.16\%. In contrast, the attacker's chances of success increase to 41.69\% and 56.45\% under EC based and greedy defense, respectively. 
For r2000 AD graphs, our results show that under independent distribution, the attacker chances of success are minimum, i.e., 25.09\% when the defender uses C-EDO based policy; the success rate increases to 28.13\% and 33.43\% under EC based defense and Greedy defense, respectively. Our proposed approach is scalable to r4000 AD graph. The results on r4000 AD graphs show that under a positive correlation, the attacker success rate is minimum, i.e., 17.29\% under C-EDO based defence; however, the success rate increases to 21.87\% and 24.18\% under EC based and Greedy defense, respectively. 

Overall, our results demonstrate that for all three AD graphs, i.e., r1000, r2000 and r4000, on average C-EDO based defence is the best defense where the attacker success rate is minimum. Also, EC based defense outperforms Greedy defense. \textit{Notably, NNDP-EDO approach \cite{goel2022defending} is not scalable to large AD graphs such as r4000 graphs; however, our proposed RL+C-EDO approach is scalable to r4000 graphs.}

\begin{table*}[t!] 
\caption{Comparison of attacker's chances of success under various attacker's policies (larger number represents better performance). Results show that the proposed RL based attacker policy approximates
the attacker’s problem more accurately than NNDP policy \cite{goel2022defending}.}
\label{r500}
\renewcommand{\arraystretch}{1.1}
\centering 
\begin{tabular}{p{2cm}p{3cm}p{2cm}p{2cm}p{1.5cm}p{1.5cm}} \hlineB{2} 

\textbf{Graph} & \textbf{Policy} &\textbf{Independent} & \textbf{Positive} & \textbf{Negative} & \textbf{Average}\\ \hlineB{2} 

r500 & RL (Proposed) & \textbf{88.01\%} & \textbf{86.58\%} &  \textbf{89.37\%}& \textbf{87.98\%}\\
 & NNDP & 87.57\% & 86.08\% & 89.28\%  & 87.64\%  \\ \hlineB{2}
r1000 & RL (Proposed) & \textbf{54.99\%} & 48.21\%$^{*}$ & \textbf{52.69\%} & \textbf{51.96\%}\\
 & NNDP & 53.52\% & \textbf{48.32\%} & 52.15\%  & 51.33\%  \\ \hlineB{2} 
r2000 & RL (Proposed) & \textbf{45.28\%}$^{*}$ & \textbf{56.41\%} & \textbf{42.43\%}$^{*}$  & \textbf{48.04\%}\\
 & NNDP & 45.11\% & 56.29\% & 42.39\%  & 47.93\%\\ \hlineB{2} 
\end{tabular} 
\end{table*}

\subsection{Experimental Setup 2}
In this experimental setup, we determine the effectiveness of our proposed RL based attacker's policy. 

\subsubsection{\textbf{Baseline}}
We compare our proposed RL based attacker's strategy with the Neural Network based Dynamic Program (NNDP) attacker policy \cite{goel2022defending}. In NNDP approach, the authors trained neural network to address the attacker's problem.

\subsubsection{\textbf{Results}}
Our baseline NNDP approach is only scalable to r2000 graph; therefore, we perform experiments on $r500$, $r1000$ and $r2000$ AD graphs. We randomly generate $10$ environmental configurations for each AD graph. 
We first train NNDP based attacker's strategy on $10$ environments for 2000 epochs and perform Monte Carlo simulations for $5000$ runs to compute the attacker's success rate on each environment. We then train our proposed RL based attacker's policy on the same set of $10$ environments for 150 epochs and then evaluate the trained policy for $5000$ episodes to compute attacker's chances of success. We reported attacker's average chances of success over $10$ environments in Table \ref{r500}. For a given environmental configuration, the attacker's policy that results in a higher success rate indicates that the corresponding policy is able to approximate the attacker's problem more accurately than others. Table \ref{r500} shows that for r500 graph, the attacker average success rate is 87.98\% under the RL policy, which is slightly higher than NNDP based policy. For the r1000 graph, attacker's average chance of success is 51.96\%, which is again higher than the NNDP policy. Our results show that the under our proposed RL based strategy, the attacker success rate is higher compared to NNDP based strategy, implying that RL policy is more effective at countering defense than NNDP policy.

\begin{table*}[t!] 
\caption{Comparison of attacker's chances of success on best defense from various attacker-defender policies (smaller number represents better performance). Results show that RL+C-EDO policy generates better defense than NNDP-EDO \cite{goel2022defending}.}
\label{e3_setting}
\renewcommand{\arraystretch}{1.1}
\centering 
\begin{tabular}{llllll} \hlineB{2} 

\textbf{Graph} & \textbf{Policy} &\textbf{Independent} & \textbf{Positive} & \textbf{Negative} & \textbf{Average}\\ \hlineB{2} 
r1000 & Best defense from RL+C-EDO (Proposed) & \textbf{40.16\%} & \textbf{41.36\%} & \textbf{41.51\%} & \textbf{ 41.01\%}\\
 & Best defense from NNDP-EDO & 42.02\% & 44.76\% & 41.53\%$^{*}$ & 42.77\%  \\ \hlineB{2} 
r2000 & Best defense from RL+C-EDO (Proposed) & \textbf{25.09\%} & 32.45\% &  \textbf{30.44\%} & \textbf{29.32\%}\\
 & Best defense from NNDP-EDO & 30.31\% & \textbf{30.17\%}$^{*}$ & 30.85\%  & 30.44\%  \\ \hlineB{2} 
\end{tabular} 
\end{table*}

\subsection{Experimental Setup 3}
In this experimental setup, we determine the effectiveness of our proposed C-EDO based defense. 

\subsubsection{\textbf{Baseline}}
We compare our proposed approach's final environmental configuration (defensive plan) with the final configuration from NNDP-EDO approach \cite{goel2022defending}. 

\subsubsection{\textbf{Results}} We run RL+C-EDO and NNDP-EDO approaches on 5 seeds from $0$ to $4$ to obtain the defender's best environment. We train the RL attacker policy for 150 epochs on the best environmental configurations from both approaches (on 5 seeds). We then evaluate the trained policy for 5000 episodes to compute the attacker success rate. We reported the results in Table \ref{e3_setting}. An environmental configuration against which the RL based attacker policy is able to achieve a lower success rate is considered as the best environmental configuration. Results in Table \ref{e3_setting} show that the average attacker success rate for r1000 AD graph is 42.77\% on the best configuration from NNDP-EDO. However, the attacker success rate is 41.01\% on the best configuration from RL+C-EDO, which is 1.76\%  less than the former approach. Similarly, for r2000 AD graph, the attacker success rate is minimal under RL+C-EDO based defensive plan. The results demonstrate that our approach RL+C-EDO is able to generate better environmental configurations and minimizes the attacker's success rate.

\subsection{Discussion}
Our first experimental setup proved that our proposed diversity based C-EDO defensive policy is better than EC based defense and greedy defense. The results also showed that EC based defense performs better than greedy defense. Our second experimental setup proved that the proposed RL based attacker policy approximates the attacker's problem more accurately than NNDP policy. For the same environment configuration, the RL based attacker's policy achieves a higher success rate than the NNDP policy. Our third experimental setup proves that the proposed  RL+C-EDO approach generates better environment configurations than NNDP-EDO, in turn minimizing the attacker's success rate. This way, our proposed approach is able to achieve the attacker's and defender's goals better than the existing approach.   Notably, our proposed approach is scalable to large-scale $r4000$ AD graphs, for which NNDP-EDO approach failed to scale. \textit{In summary, our proposed approach is highly effective, more scalable, approximates the attacker's problem more accurately and generates better defensive plans than the existing approach.}

\section{Conclusion}
We studied a Stackelberg game model in a configurable environment, where the attacker's goal is to devise a strategy to maximize their achievable rewards. The defender seeks to identify the environment configuration where the attacker’s attainable reward is minimum. We proposed a reinforcement learning based approach to address the attacker problem and critic network assisted evolutionary diversity optimization based policy to address the defender problem. We trained the attacker policy against numerous environments simultaneously. We leverage the trained RL critic network to evaluate the fitness of the environment configurations. Our experimental results showed that the proposed approach is highly effective and scalable to large-scale AD attack graphs.

\section*{Acknowledgements}
This work has been supported by  Australian Research Council through grants DP190103894 and FT200100536, and with supercomputing resources provided by the Phoenix HPC service at the University of Adelaide.

\bibliographystyle{unsrtnat}
\bibliography{references}
\end{document}